\documentclass[letterpaper, 10 pt, conference]{ieeeconf}
\IEEEoverridecommandlockouts  

\usepackage{cite}
\usepackage{amsmath,amssymb,amsfonts}
\usepackage{algorithmic}
\usepackage{graphicx}
\usepackage{textcomp}
\usepackage{xcolor}
\usepackage{amsmath}
\usepackage{amssymb}
\usepackage{mathtools}

\usepackage{amsthm}
\usepackage{algorithm}
\usepackage{algorithmic}
\usepackage{hyperref}
\usepackage[utf8]{inputenc} 
\usepackage[T1]{fontenc}    
\usepackage{hyperref}       
\usepackage{url}            
\usepackage{booktabs}       
\usepackage{amsfonts}       
\usepackage{nicefrac}       
\usepackage{microtype}      
\usepackage{xcolor}         
\usepackage{microtype}
\usepackage{graphicx}
\usepackage{subfigure}
\usepackage{makecell}
\usepackage{authblk}

     \PassOptionsToPackage{numbers, compress}{natbib}
\def\BibTeX{{\rm B\kern-.05em{\sc i\kern-.025em b}\kern-.08em
    T\kern-.1667em\lower.7ex\hbox{E}\kern-.125emX}}
\begin{document}




\title{Learning Dual-arm Object Rearrangement for Cartesian Robots}
\author{Shishun Zhang$^{1}$, Qijin She$^{1}$, Wenhao Li$^{1}$, Chenyang Zhu$^{1}$, Yongjun Wang$^{1}$, Ruizhen Hu$^{2}$, Kai Xu$^{1,3,*}$ \\
$^{1}$National University of Defense Technology $\quad$ $^{2}$Shenzhen University $\quad$ $^{3}$Xiangjiang Laboratory \\
$^{*}$Corresponding Author}

\maketitle

\begin{abstract}
	This work focuses on the dual-arm object rearrangement problem abstracted from a realistic industrial scenario of Cartesian robots. The goal of this problem is to transfer all the objects from sources to targets with the minimum total completion time. To achieve the goal, the core idea is to develop an effective object-to-arm task assignment strategy for minimizing the cumulative task execution time and maximizing the dual-arm cooperation efficiency. One of the difficulties in the task assignment is the scalability problem. As the number of objects increases, the computation time of traditional offline-search-based methods grows strongly for computational complexity. Encouraged by the adaptability of reinforcement learning (RL) in long-sequence task decisions, we propose an online task assignment decision method based on RL, and the computation time of our method only increases linearly with the number of objects. Further, we design an attention-based network to model the dependencies between the input states during the whole task execution process to help find the most reasonable object-to-arm correspondence in each task assignment round. In the experimental part, we adapt some search-based methods to this specific setting and compare our method with them. Experimental result shows that our approach achieves outperformance over search-based methods in total execution time and computational efficiency, and also verifies the generalization of our method to different numbers of objects. In addition, we show the effectiveness of our method deployed on the real robot in the supplementary video. 
\end{abstract}

\section{Introduction}
The object rearrangement problem is commonly encountered in industrial scenarios (product sorting e.g.) and has been studied in the field of task and motion planning (TAMP) \cite{beuke2018online,garrett2021integrated}. The goal of rearrangement is to transfer every object (mostly using robotic arms) from the source pick position to the target place position with the fewest total completion steps (also known as \textbf{makespan}). Utilizing multiple arms can enhance efficiency and accelerate the achievement of the goal, but introduces new higher-level task assignment challenges and lower-level motion planning challenges.

\begin{figure}
	\centering
	\subfigure[Simulation of the scene]{
		\includegraphics[scale=0.085]{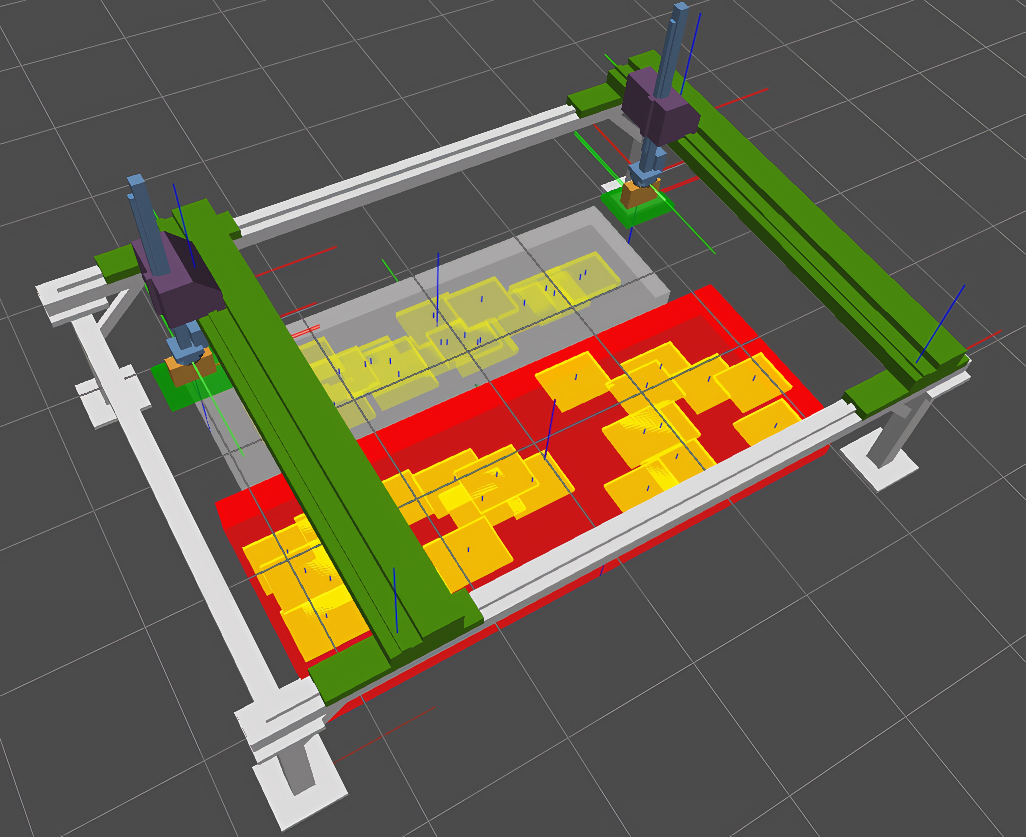}
		%
		\label{fig1:a}		
	} 
	\subfigure[Abstraction of the scene]{
		\includegraphics[scale=0.19]{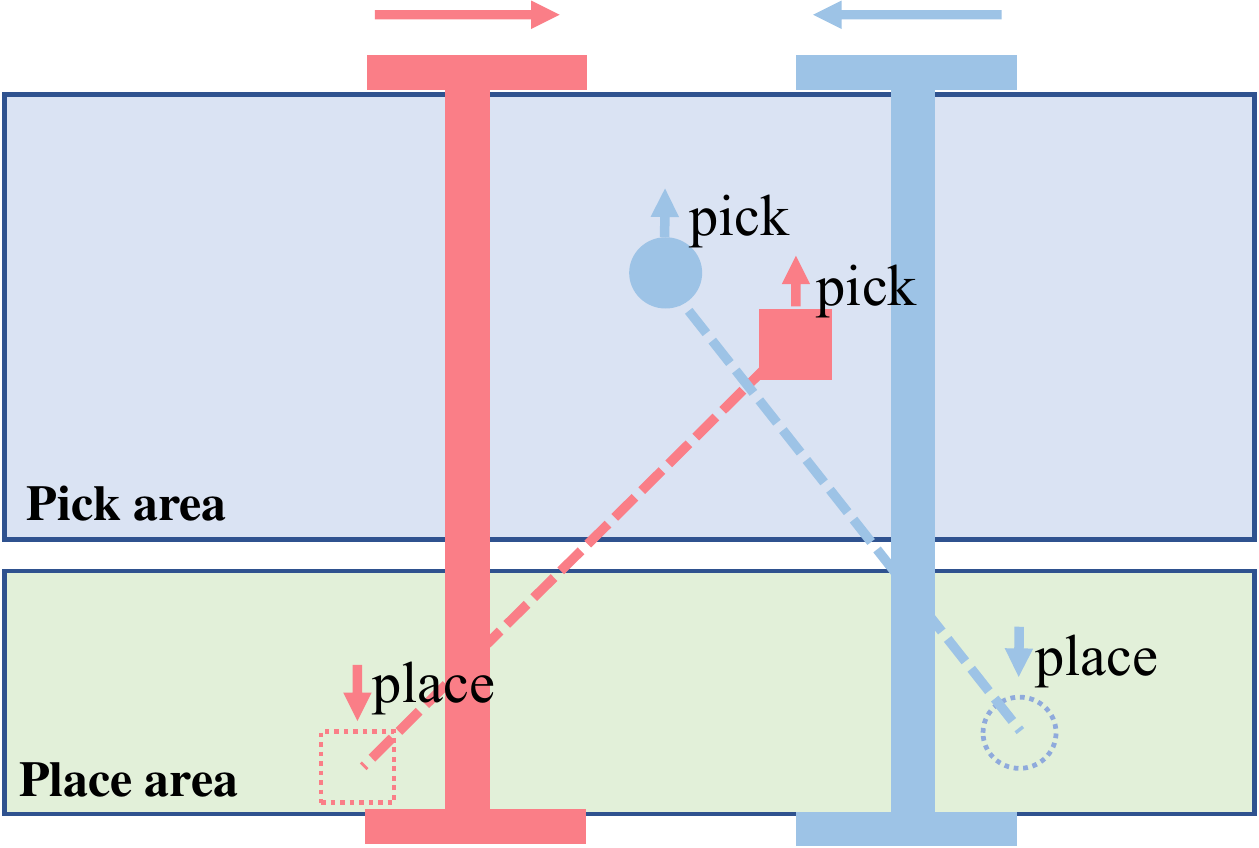}
		\label{fig1:b}
	}
	\caption{(a) Two arms in a Cartesian robot are used to transfer the objects (the yellow boards) from the red pick area to the target positions in the white place area. (b) A common motion interference situation: two arms meet, arm $r_1$ (red) and $r_2$ (blue) are about to pick up the assigned objects (the same color as the arms). To avoid the collision, $r_1$ should make a concession for $r_2$ or vice versa, which inevitably brings time delays.} 
	\label{fig_hangjia}
\end{figure}

\begin{figure}[t]
	\centering
	\subfigure[Unwise assignment: motion interference exists in the 2\textsuperscript{nd} round]{
		\includegraphics[scale=0.16]{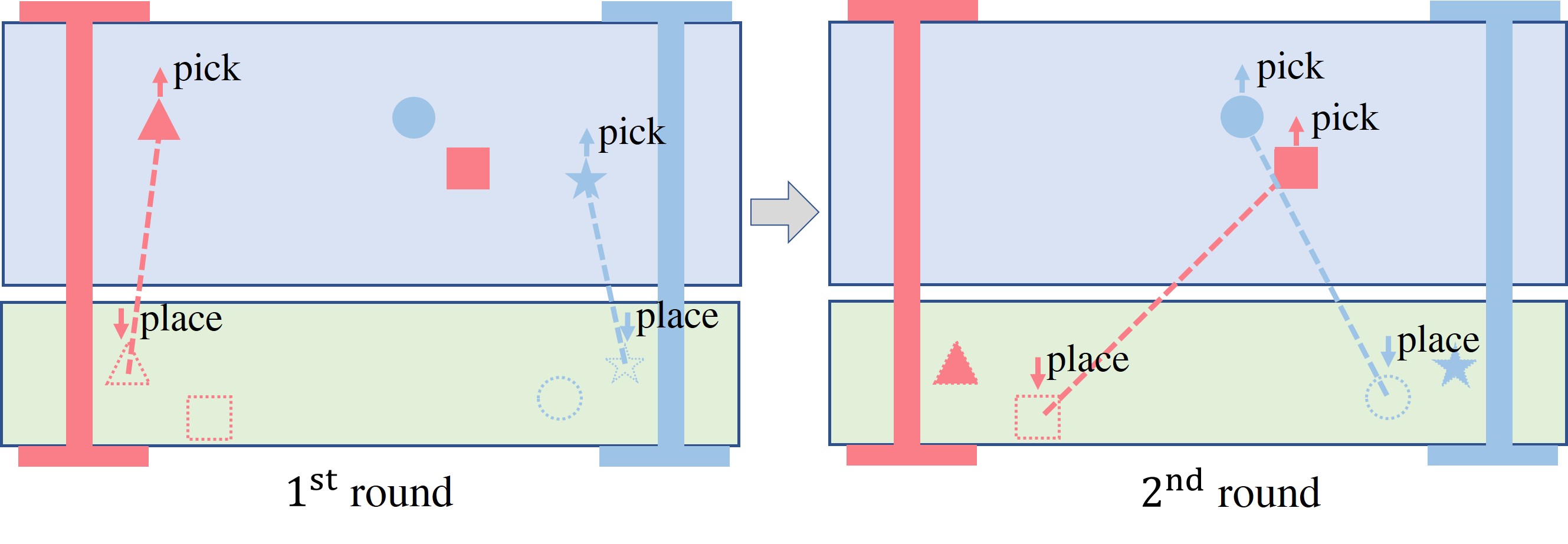}
		%
		\label{fig_assignment:a}	}	
	\subfigure[Wise assignment: no motion interference in each round]{
		\includegraphics[scale=0.16]{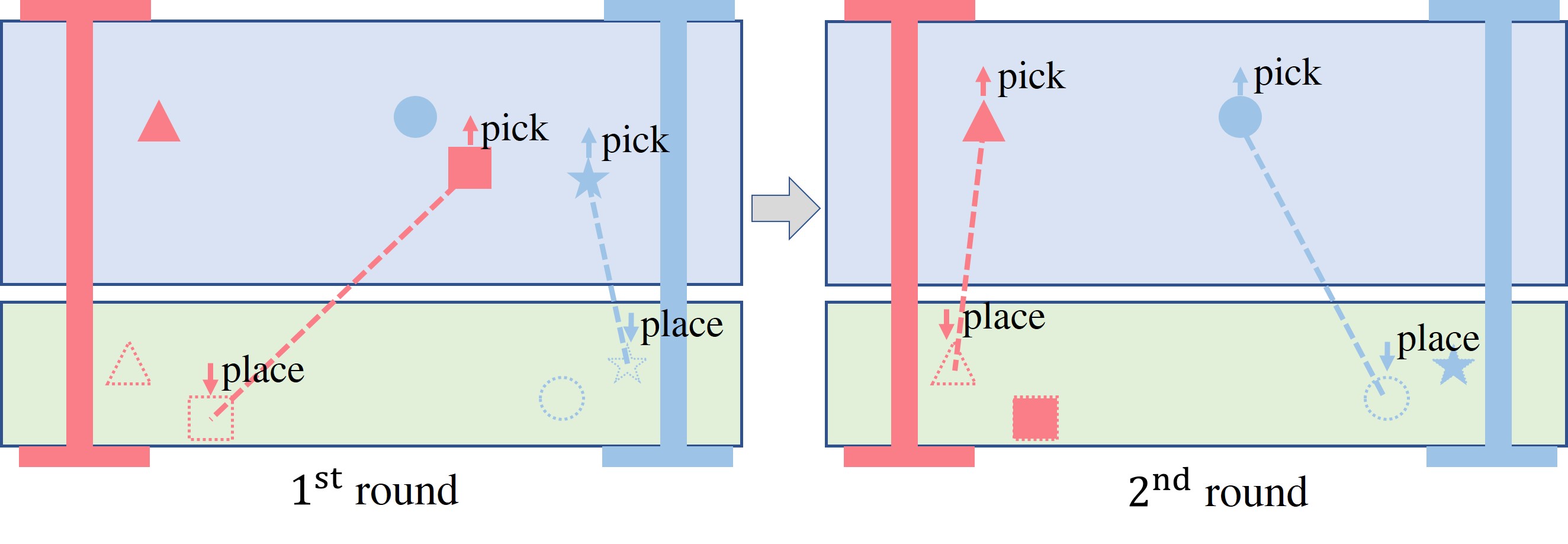}
		\label{fig_assignment:b}	}
	\caption{Comparison of the wise and unwise task assignments. (a) If both arms are assigned the object closest to them in the 1\textsuperscript{st} round, then there will be motion interference in the 2\textsuperscript{nd} round, which will bring time delays caused by concession. (b) In this way, there will be no delay either in the 1\textsuperscript{st} round or in the 2\textsuperscript{nd} round, which will shorten the total completion time.} 
	\label{fig_assign}
\end{figure}

This research primarily addresses the task assignment challenges in the dual-arm object rearrangement while also considering the actual motion planning aspect. To the best of our knowledge, this work is the first one that focuses on object rearrangement using a Cartesian robot (as shown in Figure \ref{fig1:a}). Compared with high-DOF robots, the movement of the two arms is in a relatively more limited shared space. For example, when there is an interference between the motion trajectories of the two arms (as shown in Figure \ref{fig1:b}), it cannot be avoided through trajectory optimization, such as one arm circle around another like in normal dual-arm settings, the only way is to let one arm perform the task first and another arm make a concession, which inevitably adds time delays. This limitation requires us to focus more on the higher-level part and aim to reduce the time delays by making a wise task assignment strategy as shown in Figure \ref{fig_assign}.

Assigning tasks becomes relatively straightforward when the number of objects is small. However, as the number of objects increases, the complexity of the task assignment problem grows exponentially. This poses challenges in terms of computational efficiency for traditional offline-search-based heuristic methods. Moreover, the potential dynamic changes within the scene, such as the changing positions of objects, require the offline search method to replan the entire task assignment, further adding to the inefficiency.

To address the limitations of traditional methods, we propose to make an online task assignment decision strategy. Taking inspiration from the extensive work on deep learning in online task planning \cite{mazyavkina2021reinforcement,park2023learn,paul2022learning,zhao2021online,zhao2022learning}, we recognize the potential of learning-based methods to autonomously discover effective heuristics for optimization tasks \cite{bello2016neural}. Furthermore, the efficiency demonstrated by reinforcement learning (RL) in long-sequence decision-making motivates us to adopt an online RL framework for making task assignment decisions. Another core insight of this work is utilizing the attention mechanism \cite{vaswani2017attention} to discover and model the dependencies between different kinds of input states and find the most likely object-to-arm correspondence under the guide of the dependencies. We have also designed a safe dual-arm motion planning method as the lower-level part, based on it, we adapt some search-based methods to this specific setting as the higher-level task assignment methods and compare our method with them. The experimental results demonstrate the outperformance of our method over other methods in the total completion time (makespan), and verify the acceptable complexity of our method, as it increases linearly with the number of objects.

The contribution of this research can be summarized as 1) A learning-based framework to solve the object rearrangement problems for the motion-limited dual-arm Cartesian robot. 2) An online RL method to make computationally efficient and scalable task assignment strategies for multiple objects. 3) An attention-based neural network to discover dependencies between different states and help achieve better performance in the long-sequence decision-making process.

\section{Related Works}

\textbf{Object rearrangement.} Numerous efforts have been made to address the issue of object rearrangement. Certain studies concentrate on rearrangement planning when the desired object configuration is provided \cite{king2016rearrangement,krontiris2016efficiently}. Others tackle the challenge of retrieving the target objects and relocating the surrounding objects \cite{cheong2020relocate,nam2020fast,danielczuk2019mechanical}. Furthermore, there are investigations into large-scale multiple object rearrangements \cite{huang2019large,song2020multi}. Nevertheless, these approaches primarily center around a single robot and do not incorporate multi-robot collaborative planning. In contrast, our research focuses on dual-arm collaboration in the context of object rearrangement.


\textbf{Multi-agent routing problems.} The multi-agent routing or scheduling problem, known for being NP-hard, has been extensively investigated in the field of combinatorial optimization \cite{bello2016neural,cao2022dan,khalil2017learning}. This problem can also be framed as a multiple traveling salesman problem (mTSP), for which learning-based methods have been widely employed \cite{mazyavkina2021reinforcement,park2023learn,paul2022learning,li2020towards}. However, in many of these studies, the geometric aspects and specific motions of the agents are either overlooked or idealized, despite their significance in real robot scheduling \cite{pan2021general}. Similarly, our approach treats the dual-arm rearrangement as a multi-agent routing problem, but unlike previous research, we acknowledge the realistic factors involved, by considering the actual motion of the arm, including factors such as motion interference-induced delays, as illustrated in Figure \ref{fig1:b}.


\textbf{Task assignment and collaborative planning.} Additionally, there are existing works that consider both task assignment and collaborative motion planning \cite{gao2022toward,shome2020fast, umay2019integrated}. Many search methods formulate this problem as Mixed-Integer Linear Programs (MILP) or decompose it into sub-problems to mitigate computational complexity. However, as the number of objects increases, the search time escalates significantly. Furthermore, when the number or positions of objects change, a substantial amount of time is required for replanning. In contrast, we propose an online Reinforcement Learning (RL) method that exhibits computation time linearly correlated with the number of objects.

\section{Problem Statement}

To achieve the objective of transferring all objects with minimal total execution time steps (makespan), our primary focus is on the higher-level task assignment problem, while considering the lower-level dual-arm motion planning. Here, we outline the input and output specifications of the problem:

\textbf{Input.} Suppose that there are $n$ objects $\mathcal{O}=\left\{o_{1}, \cdots, o_{n}\right\}$, we denote the source pick position and target place position of the object $o_{i}$ as $pick_{i}$ and $place_{i}$ ($i \in [1,n]$). There are two Cartesian robotic arms $\mathcal{R}=\left\{r_{1}, r_{2}\right\}$, we denote the end effector position of arm $r_{k}$ as ${ee}_{k}$ ($k \in \left\{1, 2\right\} $). 

\textbf{Output.} 1) Higher-level output: the index $i$ ($i \in [1,n]$) of the object assigned to arm $r_{k}$ ($k \in \left\{1, 2\right\} $) at every task assignment round $\tau$ (we use $A^k_{\tau}=i$ to indicate that). 2) Lower-level output: the collision-free dual-arm motion trajectory when a task is assigned at every $\tau$.

\textbf{Task assignment.} In our real robot system, the control host synchronously publishes the assigned tasks for both arms. Only when both arms have completed their current tasks, the host will publish the new task pair.  We denote the set of assignment pairs as $\mathcal{A}$ and the assignment pair at each round $\tau$ as $\mathcal{A}_{\tau}$, we have:
\begin{equation}\label{eq_3_1}
	\begin{aligned}
		\mathcal{A}_{\tau}=\left\{A^1_{\tau},A^2_{\tau}\right\}_{\tau=1, \cdots, T}.
	\end{aligned}
\end{equation}
We let $A^1_{\tau}=i, A^2_{\tau}=j$ ($i \ne j; i,j \in [1,n]$) represent the index of the object assigned to $r_{1}$ and $r_{2}$ respectively. 
For each assignment pair $\mathcal{A}_{\tau}$, we denote $m_{\tau}$ as the number of time steps required to complete the assignment pair. We need to find the optimal object-to-arm assignment $\mathcal{A^*}$ to minimize the makespan of the whole rearrangement process:
\begin{equation}\label{eq_assignment_goal}
	\begin{aligned}
		\mathcal{A^*}=\underset{\mathcal{A}}{\operatorname{argmin}} \sum_{\tau=1}^{T} m_{\tau}.
	\end{aligned}
\end{equation}

\textbf{Dual-arm motion planning.} When the assignment pair $\mathcal{A}_{\tau}$ is determined, there should be a collision-free dual-arm motion trajectory to complete the transferring process, we denote as $M(\mathcal{A}_{\tau})$. Due to the actual discrete motion control, $M(\mathcal{A}_{\tau})$ consists of the discrete actions of each arms:
\begin{equation}\label{eq_3_1_1}
	\begin{aligned}
		M(\mathcal{A}_{\tau})= \left\{M^1_{\tau},M^2_{\tau}\right\}_{\tau=1, \cdots, T}.
	\end{aligned}
\end{equation}
Where $M^1_{\tau}$ and $M^2_{\tau}$ are the sets of the discrete steps of arm $r_1$ and $r_2$ respectively. As mentioned above, there should be no collision between $M^1_{\tau}$ and $M^2_{\tau}$ at every motion step.

\section{Method}
\label{section4}

We primarily concentrate on addressing the higher-level task assignment problem and present a learning-based framework to achieve it. Additionally, we have developed a heuristic-based approach to execute collision-free dual-arm cooperative motion planning.

\subsection{Higher-level task assignment}

Drawing inspiration from previous studies on learning-based approaches for large-scale routing problems \cite{cao2022dan,gao2023amarl,park2023learn}, and considering the adaptability of online reinforcement learning (RL) in long-sequence decision-making, we model task assignment as a Markov decision process. We then incorporate it into a deep RL framework, treating the two arms as separate agents. We interpret the task assignment process as an active dual-arm task selection process.


\textbf{MDP formulation.} We formulate the online task assignment process as a typical discrete-time Markov decision process, which can be described as a tuple of $(\mathcal{S}, \mathcal{A}, \mathcal{P}, \mathcal{R}, \gamma)$, where $\mathcal{S}$ is the set of environment states; $\mathcal{A}$ is the action set, which corresponds to the assignment process in our problem; $\mathcal{R}: \mathcal{S} \times \mathcal{A} \rightarrow \mathbb{R}$ is the reward function; $\mathcal{P}: \mathcal{S} \times \mathcal{A} \times \mathcal{S} \rightarrow \left[0,1 \right]$ is the transition probability function; $\gamma$ is the discount factor. The policy $\pi: \mathcal{S} \rightarrow \mathcal{A}$ is a map from states to probability distributions over actions. We seek for a policy $\pi$ to maximize the accumulated discounted reward:
\begin{equation}\label{discounted reward}
	\begin{aligned}
		J(\pi)=E_{\pi}\left[\sum_{\tau=1}^{T} \gamma^\tau \mathcal{R}\left(s_\tau, a_\tau\right)\right] .
	\end{aligned}
\end{equation}
Here $\tau$ corresponds to the assignment round in our problem, and the action $a_\tau$ is the assignment pair $\mathcal{A}_{\tau}$ that is decided by the policy network at each step.

\textbf{Observation.} The observation consists of three parts: arm states, object states, and a global mask. The arm state $s_{k}^{r}=\left(x^{r}_{k}, y^{r}_{k} \right), k \in\{1, 2\}$ contains the dynamic 2-$d$ end-effector coordinates of all arms; The object state $s_{i}^{o}=\left(x_{i}^{s}, y_{i}^{s}, x_{i}^{t}, y_{i}^{t} \right), i \in [1,n]$ contains the source pick and target place position of all objects, and we also update a global mask $M=\{M_1, \cdots, M_n \}$. Each $M_i$ is initially 1, and set to 0 if any object $o_i$ has been transferred, which indicates that the object cannot be assigned in the next round.

\textbf{Action.} At each assignment round $\tau$, based on the current observation $\left(s^{r}_{\tau}, s^{o}_{\tau}, M \right)$, our network outputs a policy $\pi_{\theta}\left(\mathcal{A}_{\tau} \right)$, parameterized by the set of weights $\theta$:
\begin{equation}\label{eq_action}
	\begin{aligned}
		\pi_{\theta}\left(\mathcal{A}_{\tau}=\left(i,j \right) \mid s^{r}_{\tau}, s^{o}_{\tau}, M \right) , i,j \in [1,n].
	\end{aligned}
\end{equation}
Where $i,j$ indicate the object index assigned to $r_1$ and $r_2$.

\textbf{Reward Structure.} Since our goal is to minimize the makespan (the total completion time) of the whole process (as illustrated in Eq. \ref{eq_assignment_goal}), we follow the most direct and general design, and use the minus of makespan as the sparse reward:
\begin{equation}\label{eq_reward}
	\begin{aligned}
		R(\pi)= - \sum_{\tau=1}^{T} m_{\tau}.
	\end{aligned}
\end{equation}
Where $m_{\tau}$ is mentioned above as the number of time steps required in one assignment round. 

\subsection{Attention-based neural network}
\label{Attention-based neural networks}
\begin{figure}[t]	
	\centering
	\includegraphics[scale=0.4]{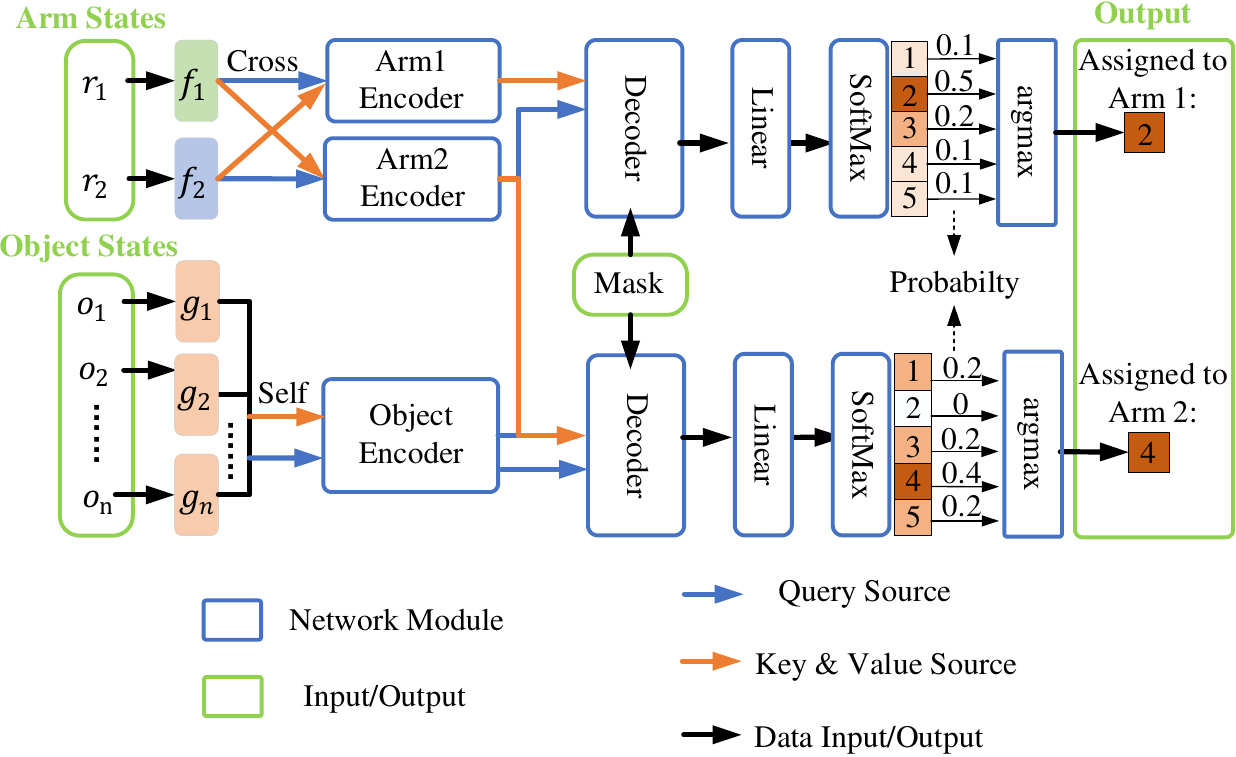}
	\caption{The structure of the attention-based task assignment network.}
	\label{fig3}	
\end{figure}
Our main contribution is a centralized attention-based neural network, illustrated in Figure \ref{fig3}, which is shared by both the policy and Q-function networks. The key idea behind our network design is to employ the attention mechanism to identify and capture the relationships between various state information, such as arm-to-arm, object-to-object, and arm-to-object dependencies. These dependencies are then utilized to facilitate the assignment decision-making process. Moreover, the attention mechanism possesses order-invariant and scale-invariant properties, enabling the network to generalize well across different numbers of objects.

\textbf{Pipeline.} The attention-based arm encoders and object encoder receive the arm states and object states as inputs. They model the dependencies among arms and among objects, respectively, and generate arm and object features as outputs. Each arm feature, combined with the shared object feature, forms a pair of (Query, Key-Value) inputs for two attention-based decoder modules. Within each decoder module, the dependency between an arm and all objects is modeled. Guided by the global mask $M$, attention weights are assigned to each object for the arm. Finally, we use the SoftMax function to output the probability and assign the object with the highest probability to each arm in a greedy manner. The detailed process is further explained in the subsequent section.

\textbf{Initial state encoder.} We use two 2-layer multilayer perceptrons (MLP) to model arm encoding function $\left\{f_{1}(\cdot),f_{2}(\cdot)\right\}$ and object encoding function $\left\{g_{1}(\cdot), g_{2}(\cdot), \cdots, g_{n}(\cdot)\right\}$. Each function takes the state of arms or objects as the input and outputs $d$-dimensional ($d=128$ in practice) arm embeddings $h_{i}^{r}$ ($i \in \left\{1, 2\right\}$) or object embeddings $h_{j}^{o}$ ($j \in [1,n]$). 

\textbf{Arm encoder.} For each arm, the initial embedding $h_{i}^{r}$ is passed through a multi-head attention layer (MHA) as the query source $h^{q}$, with the initial embedding of another arm serving as the key-and-value source $h^{k,v}$. By doing so, we achieve the cross-attention of two arm to model dependencies between them. We term the output as $ h_{i,j}^{\prime r}$ for each arm.


\textbf{Object Encoder.} Since self-attention achieved good performance to model the message passing of nodes in \cite{kool2018attention}, we rely on the same idea, and also use MHA to model the dependencies of all objects in our problem. Specifically, all the $n$ initial object embeddings $h_{j}^{o}$ are concated into a $n\times d$ tensor $h^{O}$. Then the tensor are served as both query source and key-and-value source of the attention layer, as is commonly done in self-attention mechanisms. We term the output as $ h^{\prime O}$.

\begin{figure}[t]
	\centering
	\subfigure[object number=6]{
		\includegraphics[scale=0.12]{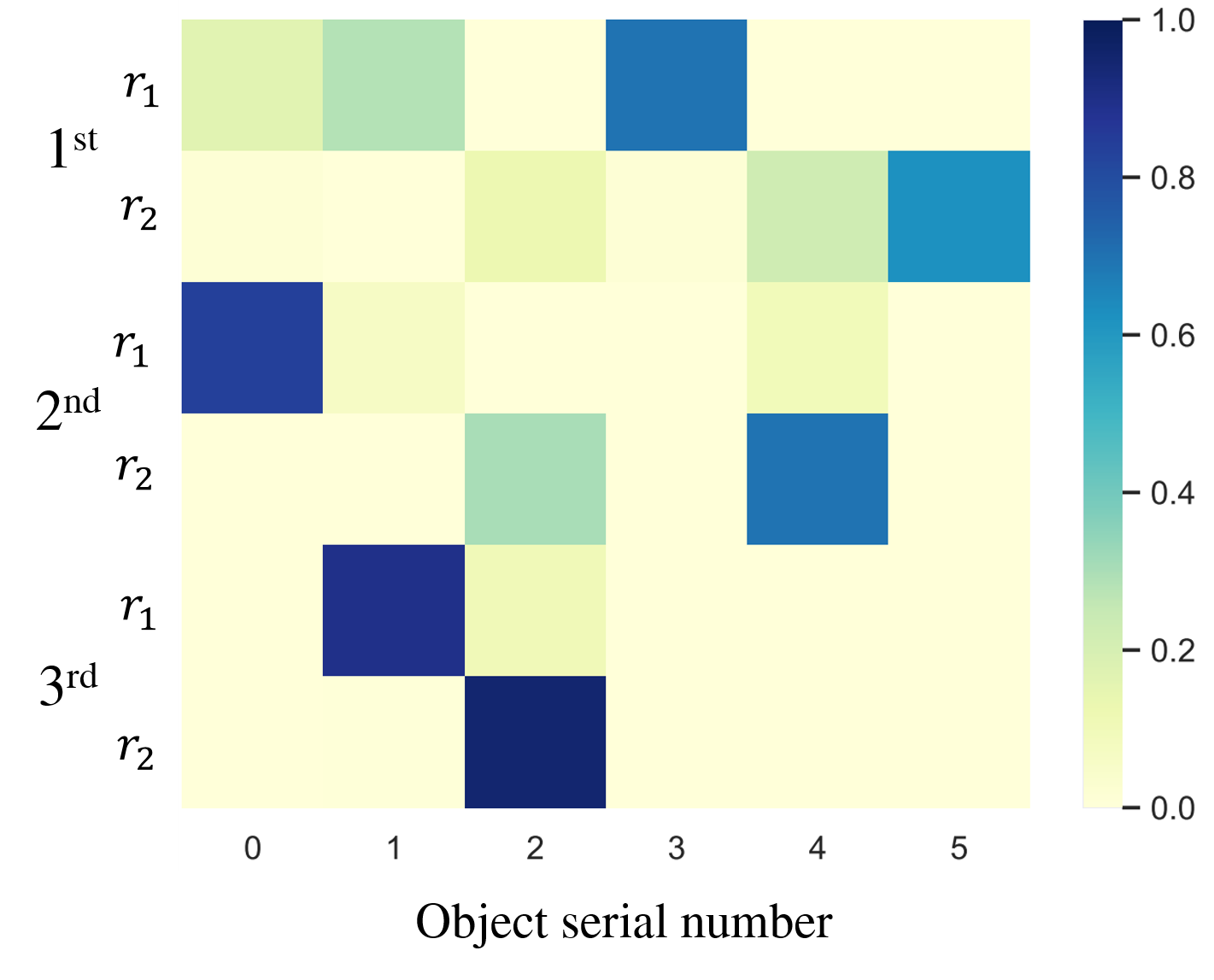}
		%
		\label{fig_attention:c}	}	
	\subfigure[object number=10]{
		\includegraphics[scale=0.16]{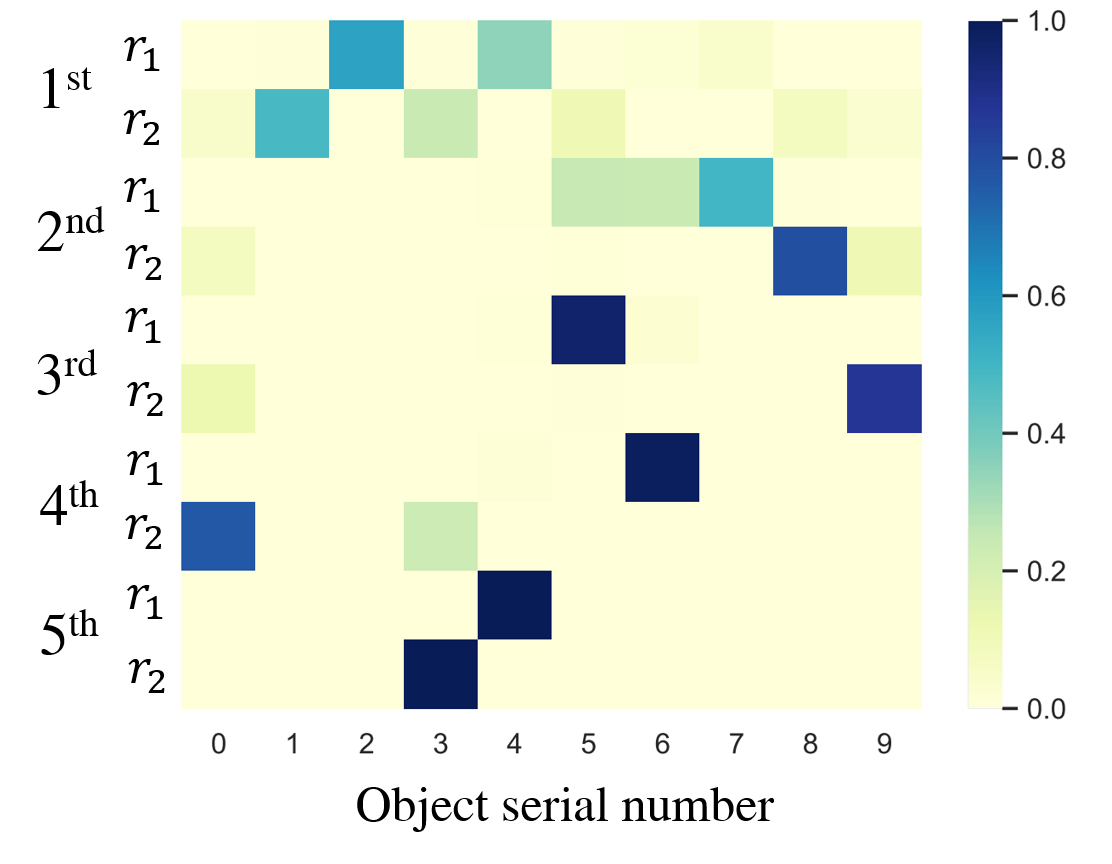}
		%
		\label{fig_attention:a}		
	} 
	\caption{Visualization of the arm-to-object attention in each assignment round. Each row corresponds to the degree of attention a robot arm paid to different objects in a certain round. (represented as “1\textsuperscript{st}”,“2\textsuperscript{nd}” round, etc.) }
	\label{fig_attention}	
\end{figure}

\textbf{Decoder.} The input of the decoder consists of the common object embedding $ h^{\prime O}$ (as the query source), an arm embedding $ h_{i}^{\prime r}, i \in\{1, 2\}$ (as the key-and-value source), and the global mask $M$. There, we rely on the global mask $M$ to manually set the similarity of each object $o_j$ to each arm $r_i$ as $u^{i}_{j}=-\infty, (i \in \left\{1, 2\right\}, j \in [1,n])$, to ensure that if the object $o_j$ has been rearranged, the attention weight is 0.
\begin{equation}\label{eq1}
	u^{i}_{j}=\left\{\begin{array}{cl}
		\frac{q^{T} \cdot k^{i}_{j}}{\sqrt{d}} & \text { if } M_{j}=1 \\
		-\infty & \text { otherwise }
	\end{array}\right.
\end{equation}
These similarities are normalized using a Softmax, to finally yield the attention map $p$ of the object $o_j$ for each arm $r_i$ in each round, as shown in Figure \ref{fig_attention}.
\begin{equation}\label{eq2}
	p^{i}_{j}=\pi^{i}_{\theta}\left(A^i_{\tau}=j \mid h^{\prime O}, h_{i}^{\prime r}, M\right)=\frac{e^{u^{i}_{j}}}{\sum_{j=1}^{n} e^{u^{i}_{j}}}.
\end{equation}
\textbf{Implement details} We represent both actor and critic with our weight-sharing network under the Proximal Policy Optimization (PPO; \cite{schulman2017proximal}) structure for its stability in training and robustness to the choice of hyperparameters. 
\subsection{Lower-level motion planning}
The lower-level component is designed to ensure safe dual-arm cooperative motion planning once the assignment pair is determined at each round $\tau$. To accomplish this, we utilize a heuristic-based method to plan a collision-free motion trajectory for both arms before executing the assigned task.

\begin{figure}[t]
	\centering
	\subfigure[Interference detection]{
	    \includegraphics[scale=0.3]{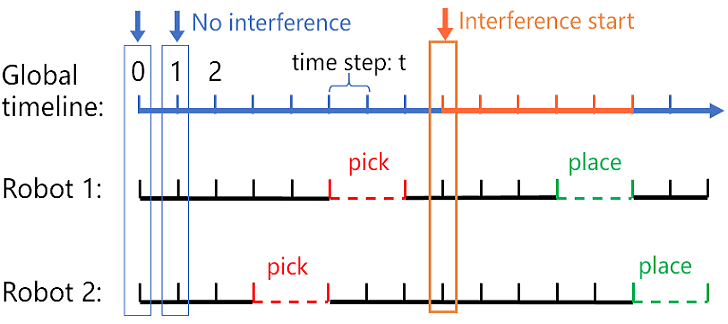}
	    \label{fig_replan:a}		}	
	\subfigure[Motion replanning]{
		\includegraphics[scale=0.18]{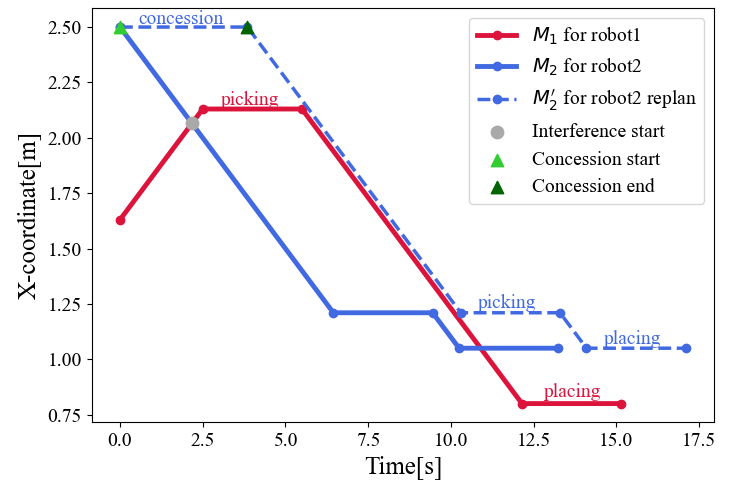}
		\label{fig_replan:b}			
	} 
	\caption{(a) Discretize the trajectories into fixed interval time steps, the solid line indicates going to the pick/place position, while the dashed line indicates picking/placing. The interference is checked at every step. (b) An example of motion replanning when interference exists. The solid lines indicate $M_1$ and $M_2$ (minus the safe distance two arms need to keep), and the dashed line indicates $M^\prime_2$, the replanned trajectory for $r_2$.}
\end{figure}

\begin{algorithm}[t]
	\renewcommand{\algorithmicrequire}{\textbf{Input:}}
	\renewcommand{\algorithmicensure}{\textbf{Output:}}
	\caption{Lower-level motion planning method}
	\label{alg:low-level algorithm}
	\begin{algorithmic}[1]
		\REQUIRE
		The source end-effector position $ee_{1}^{S}, ee_{2}^{S}$; the pick and place position $pick_{1,2}, place_{1,2}$. 
		\ENSURE
		The collision-free trajectories $M_1, M_2$; the total time step $T$; the final end-effector position $ee_{1}^{F}, ee_{2}^{F}$.
		\STATE $M_1, M_2\leftarrow$ InitPlan($ee_{1,2}^{S}, pick_{1,2}, place_{1,2}$) ; 
		\STATE $M_1, M_2 \leftarrow $ Discretize($M_1,M_2$);
		\STATE is\_interference, $t_{intf\_start}$ $\leftarrow$ CheckCollision($M_1,M_2$);\WHILE{is\_interference = True}
		\STATE $M_{1st,2nd}, t_{conc1,conc2} \leftarrow$ PriorityDecide($M_1,M_2$);
		\STATE $M_{2nd} \leftarrow$ Replan($t_{intf\_start},t_{conc1,conc2} ,M_1,M_2$);
		\STATE $M_{1st} \leftarrow M_{1st}$; // no need to replan
		\STATE is\_interference, $t_{intf}$ $\leftarrow$ CheckCollision($M_{1st},M_{2nd}$);
		\ENDWHILE
		\STATE $T_1=|M_1|$, $T_2=|M_2|$;
		\STATE $T= \max(T_1,T_2)$, $ee_{1}^{F}=M_{1}(T_{1}), ee_{2}^{F}=M_{2}(T_{2})$;
		\RETURN $M_1, M_2, T, ee_{1}^{F}, ee_{2}^{F}$ 
	\end{algorithmic}
\end{algorithm}

As illustrated in algorithm \ref{alg:low-level algorithm}, the lower-level method first plans the initial motion trajectories $M_1$ and $M_2$ for $r_1$ and $r_2$. Then both trajectories are discretely divided into fixed time interval steps, and the interference is checked at each step, as shown in Figure \ref{fig_replan:a}; If no interference, both arms will run according to the initial trajectory, and the total time steps will be returned. Otherwise, the interference start time $t_{intf\_start}$ will be recorded, then the priority of the two trajectories will be determined to help achieve the minimum concession time. The trajectory with the higher priority will be marked as $M_{1st}$, and another as $M_{2nd}$. Referring to $t_{intf\_start}$ and the movement range of $M_{1st}$, the concession start and end time $t_{conc1}, t_{conc2}$ of $M_{2nd}$ will be calculated, and the trajectory will be replanned as $M^\prime_{2nd}$. If motion interference still exists between $M_{1st}$ and $M^\prime_{2nd}$ in subsequent steps, then continue the process of check-and-replan, until there is no interference between the two trajectories. 

Figure \ref{fig_replan:b} shows an example of motion replanning in the $x$-direction (the direction of the long common rail in Figure \ref{fig1:a}) of arms that change with time. The figure simply uses straight lines as the acceleration in the real robot is large enough, that the movement process can be approximated as uniform motion at the set speed. It is also worth noting that, in actual control, the potential interference between each arm and surroundings or objects will also be checked.



\subsection{Implementation details}
\label{training}
Following the RL training method, in every episode, we randomly sample $n$ objects in the scene as the training data.

\textbf{Data generation.} We manually divide the scene into three parts according to the reachable range of the two arms on the $x$-axis. Combined with Figure \ref{fig1:b}, we define the two 1/4 size areas on the left and right sides as  “exclusive areas” for each arm, and the central 1/2 size areas as “common area” for both arms. If the pick or place position of an object is located in the exclusive area of an arm, the object can only be operated by this arm. If the pick and place positions are both located in the common area, the object can be operated by both arms. From this, we propose two sampling schemes: 1) \textbf{FS}: the pick and place positions of all objects are sampled uniformly in the full space while avoiding the case where the pick and place positions are located in two exclusive areas (to ensure each object is operable by at least one arm) 2) \textbf{CA}: all the pick and place positions are sampled only in the narrow common area, which will cause more possibilities for motion interference than \textbf{FS}, and brings greater challenges to reducing delay time through task assignment.

\textbf{Training details.} We use both sampling schemes to train our networks. When using the \textbf{FS} sampling for training, there could be some “unreachable assignments” since the policy network might assign the object that is only operable by one arm to another arm. To avoid this situation, we add an extra mask to the global mask $M$ and guarantee that the network will not assign unreachable objects to any arm.  All networks and learnable parameters are trained with Adam optimizer \cite{kingma2014adam} with a learning rate ${10}^{-3}$, $\beta_1 =0.9$, $\beta_2 =0.999$. 
\section{Experiments}
In this section, we evaluate the makespan, delay proportion, and computation time across approaches to demonstrate that our method (i) can help discover efficient long-sequence task assignment strategies that can reduce the overall delay and achieve outperformance in the makespan; (ii) can be generalized to a large number of objects when trained on a relatively small number of objects; (iii) is computationally more efficient than the search-based methods. 
\subsection{Experiment setting}

We train our networks and run all experiments with an NVIDIA 3080Ti GPU and an i9-10980XE CPU.

\textbf{Baselines.} The performance of our method is compared with three search-based methods and we have also provided two ablations of our method.

1) \textit{Perfect Matching + DP}: We reproduced the method of Shome et al. \cite{shome2020fast} and adapted it to our problem. The method leverages Perfect Matching \cite{galil1986ev} and Dynamic Programming (DP) to make offline decisions on task assignments. Based on optimal matching over an undirected graph (transfer graph), the runtime complexity of the method is $O(|\mathcal{E}||\mathcal{V}| \log |\mathcal{V}|)=O\left({ }^{n} P_{2} n \log n\right)=O\left(n^{3} \log n\right)$.

2) \textit{Random Split}: In every round of assignment, we randomly assign objects to the two arms while following the condition of no “unreachable assignment”. 

3) \textit{Greedy Search}: The “greedy” does not mean simply assigning the object closest to an arm, but assigning the task pair with the least cost currently. In every round of assignment, we calculate the operation steps of every possible assignment pair with the lower-level method and choose the assignment pair with the fewest operation steps for this round. The runtime complexity is  $O\left({ }^{n} P_{2} n^2\right)=O\left(n^{4}\right)$.

4) \textit{No\_object\_encoder}: We use the same architecture as our model, but without the object encoder, and all of the initial object embeddings are just concat together as the query source of two decoders. 

5) \textit{No\_arm\_encoder}: We use the same network model, but without arm encoders, and all of the initial arm embeddings are directly input as the sources to the two decoders. 

\textbf{Performance metrics.} To evaluate the performance of our method and all baselines, we make two test sets with the sampling scheme of \textbf{FS} and \textbf{CA}. We sample the object number = 4,6,10,14,20,30 with both schemes and each case with 1000 instances. We train our network, \textit{No\_object\_encoder} and \textit{No\_arm\_encoder} with object number = 10, and generalize the trained model to all cases. For heuristic methods, we apply them directly to all cases. 

\subsection{Results}
\begin{figure*}[t]
	\centering
	\subfigure[object number=4]{
		\includegraphics[scale=0.16]{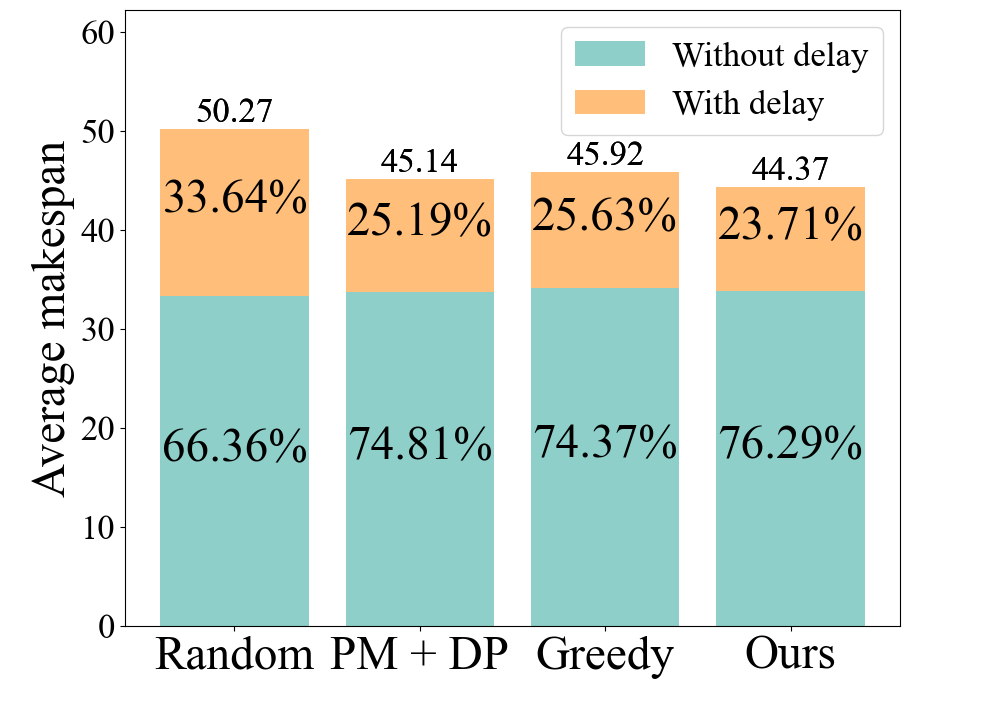}
		%
		\label{fig_delay:a}		
	} 
	\subfigure[object number=10]{
		\includegraphics[scale=0.16]{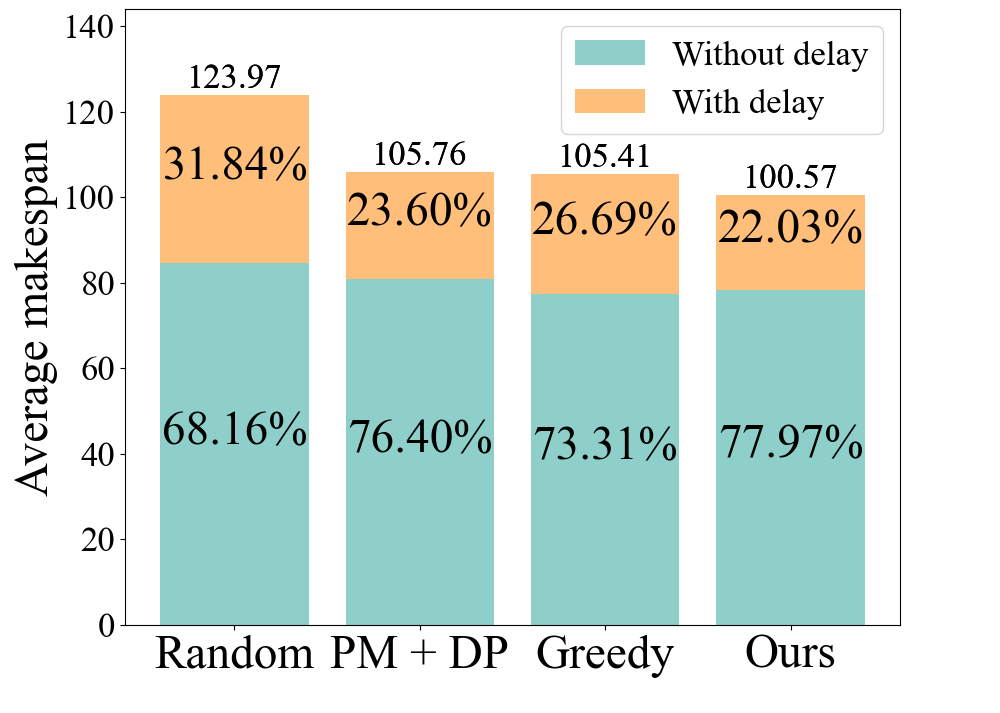}
		%
		\label{fig_delay:b}		
	}
        \subfigure[object number=20]{
		\includegraphics[scale=0.16]{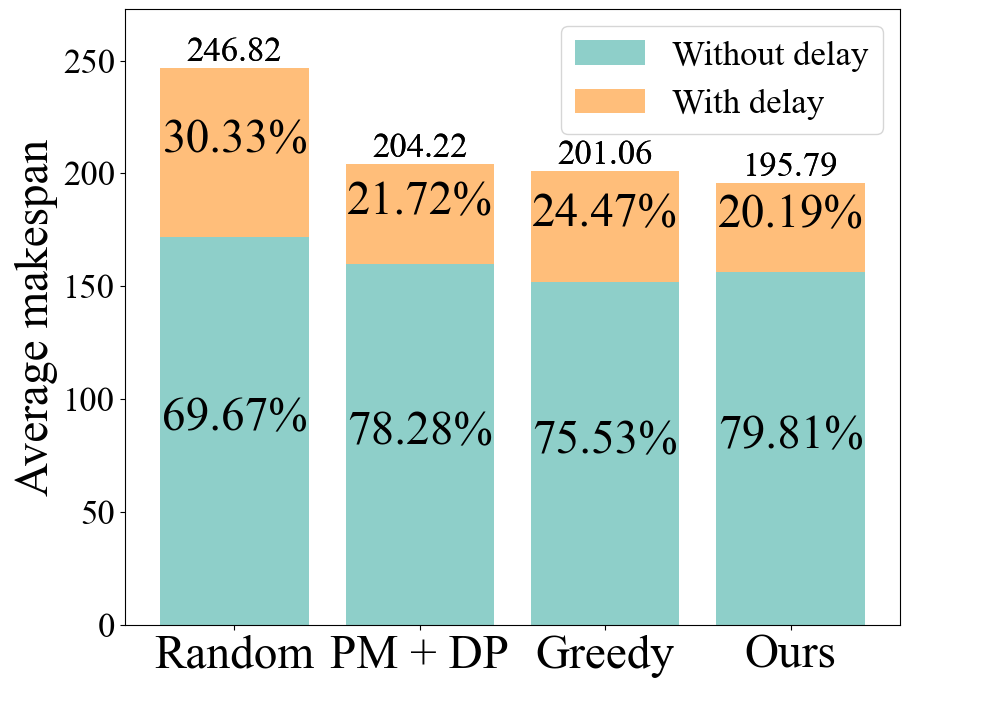}
		%
		\label{fig_delay:c}		
	}
        \subfigure[object number=30]{
		\includegraphics[scale=0.16]{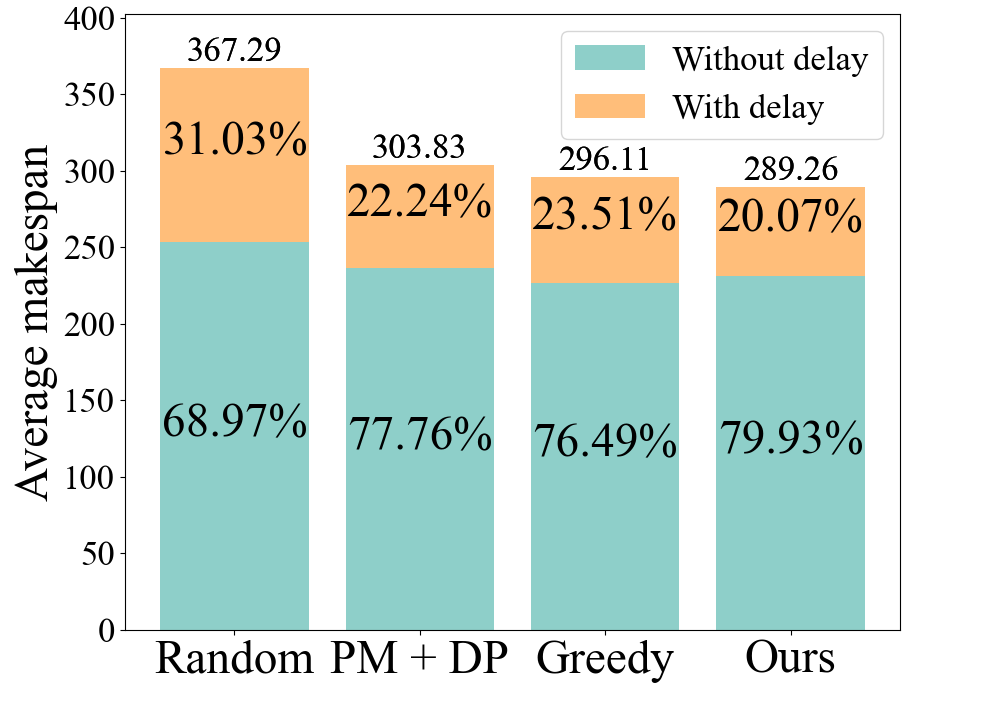}
		%
		\label{fig_delay:d}		
	} 
	\caption{The proportion of delay (or concession) time of \textit{Random Split}, \textit{Perfect Matching + DP}, \textit{Greedy Search}, and  \textit{Ours} (lower is better). “With delay” corresponds to the situation where one arm makes a concession for the other, and “Without delay” is the case of both arms at work. } 
	\label{fig_delay}
\end{figure*}

\begin{figure}[t]	
	\centering
	\includegraphics[scale=0.25]{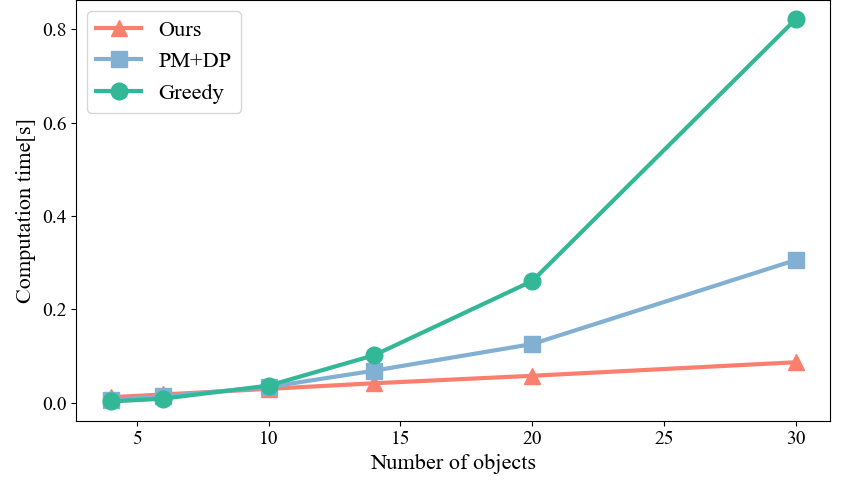}
	\caption{Comparison of the average computation time.} 
	\label{fig5}	
\end{figure}

In Table \ref{tab:1} and Table \ref{tab:2} we report the average Makespan (lower is better) of our method and all baselines on the two test sets we made. In Figure \ref{fig_delay} we present the average proportion of delay time within the total makespan for our method and the heuristic approaches, on the \textbf{CA} test set, for it sampled only from the narrow center area which significantly increases the probability of motion interference and delay occurring. In Figure \ref{fig5}, we show the comparison of average computation time executing different numbers of objects for \textit{Perfect Matching + DP}, \textit{Greedy Search}, and our method.

From the results of average makespan in Table \ref{tab:1} and Table \ref{tab:2}, we first notice that our method can well generalize to all cases despite only trained under the case of n = 10, and it is optimal in most of the cases in the two test sets, except in the case of object number = 20 and 30 in \textbf{FS} test set, in which \textit{Greedy Search} performs best. We analyze that, as object numbers increase, \textit{Greedy Search} has advantages over other methods by virtue of a large number of searches and comparisons of assignment pairs in each round. However, this advantage is a trade of computation time, as shown in Figure \ref{fig5}, the search time of which increases dramatically with respect to the scale of the instance, while our method only increases linearly. The computation time of \textit{Perfect Matching + DP} also shows polynomial time growth. From the result, we can say that our method achieves a good balance between computation efficiency and performance. 
\begin{table}
	\centering
	\caption{Results of average makespan on the test set of \textbf{CA} sampling (1000 instances each). n denotes the number of objects.}
	\label{tab:1}
	\resizebox{\columnwidth}{!}{
	\begin{tabular}{ccccccc}
		\toprule
		\multicolumn{1}{c}{Method}  & n=4 & n=6 & n=10 & n=14 & n=20 & n=30 \\
		\midrule
		\multicolumn{1}{l}{Perfect Matching + DP} & 45.14  & 65.76 & 105.76 & 145.45 & 204.22 & 303.83 \\
		\multicolumn{1}{l}{Random Split}   & 50.27 & 75.45 & 123.97 & 175.37 & 246.82 & 367.29 \\
		\multicolumn{1}{l}{Greedy Search} & 45.92  & 66.78 & 105.41 & \underline{143.70} & \underline{201.06} & \underline{296.10} \\
		\multicolumn{1}{l}{No\_object\_encoder}   & 47.34  & 67.81 & 109.45 & 147.19 & 212.42 & 315.02 \\
		\multicolumn{1}{l}{No\_arm\_encoder}   & \underline{44.93}  & \underline{64.79} & \underline{105.32} & 145.78 & 205.26 & 306.97 \\
		\multicolumn{1}{l}{Ours}   & \textbf{44.37}  & \textbf{63.13} & \textbf{100.57} & \textbf{138.69} & \textbf{195.79} & \textbf{289.26} \\
		\bottomrule
	\end{tabular}}
\end{table}
\begin{table}
	\centering
	\caption{Results of average makespan on the test set of \textbf{FS} sampling(1000 instances each).}
	\label{tab:2}
	\resizebox{\columnwidth}{!}{
	\begin{tabular}{ccccccc}
		\toprule
		\multicolumn{1}{c}{Method}  & n=4 & n=6 & n=10 & n=14 & n=20 & n=30 \\
		\midrule
		\multicolumn{1}{l}{Perfect Matching + DP} & 38.23  & 55.44 & 89.05 & 121.80 & 171.72 & 252.30 \\
		\multicolumn{1}{l}{Random Split}   & 39.65 & 57.75 & 96.75 & 136.41 & 191.67 & 285.84 \\
		\multicolumn{1}{l}{Greedy Search} & 38.22  & 55.67 & 89.13 & 121.23 & \textbf{170.08} & \textbf{248.59} \\
		\multicolumn{1}{l}{No\_object\_encoder}   & 38.17  & 55.39 & 92.80 & 125.87 & 177.20 & 259.57 \\
		\multicolumn{1}{l}{No\_arm\_encoder}   & \underline{38.12}  & \underline{55.27} & \underline{87.43} & \underline{120.51} & 170.19 & 252.09 \\
		\multicolumn{1}{l}{Ours}   & \textbf{37.84} & \textbf{54.64}  & \textbf{87.05} & \textbf{120.04} & \underline{170.15} & \underline{251.22} \\
		\bottomrule
	\end{tabular}}
\end{table}

\begin{figure}[t]
	\centering
	\subfigure[CA Sampling]{
		\includegraphics[scale=0.23]{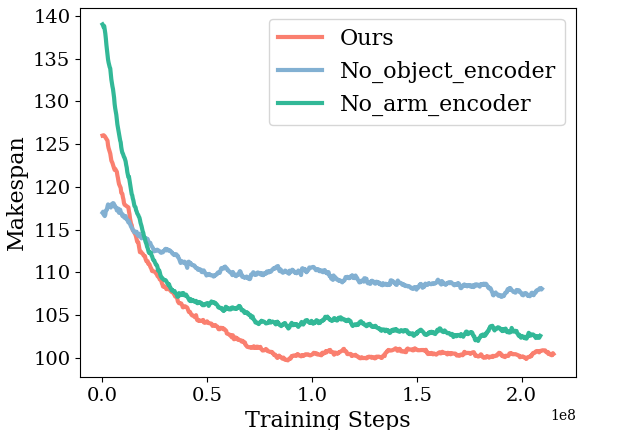}
		%
		\label{fig4:a}		
	} 
	\subfigure[FS Sampling]{
		\includegraphics[scale=0.23]{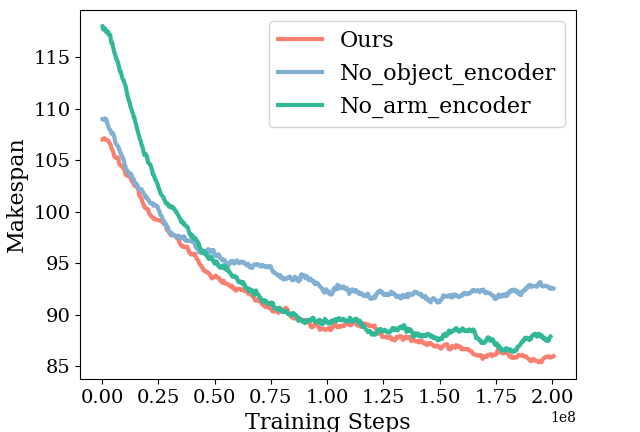}
		\label{fig4:b}
	}
	\caption{\textbf{Training ablation.} Makespan over the training steps. We smoothed all the curves with a smoothing rate of 0.98 to facilitate clearer.} 
	\label{fig4}
\end{figure}

\begin{figure}[t]
	\centering
	\subfigure[Simulator]{
		\includegraphics[scale=0.13]{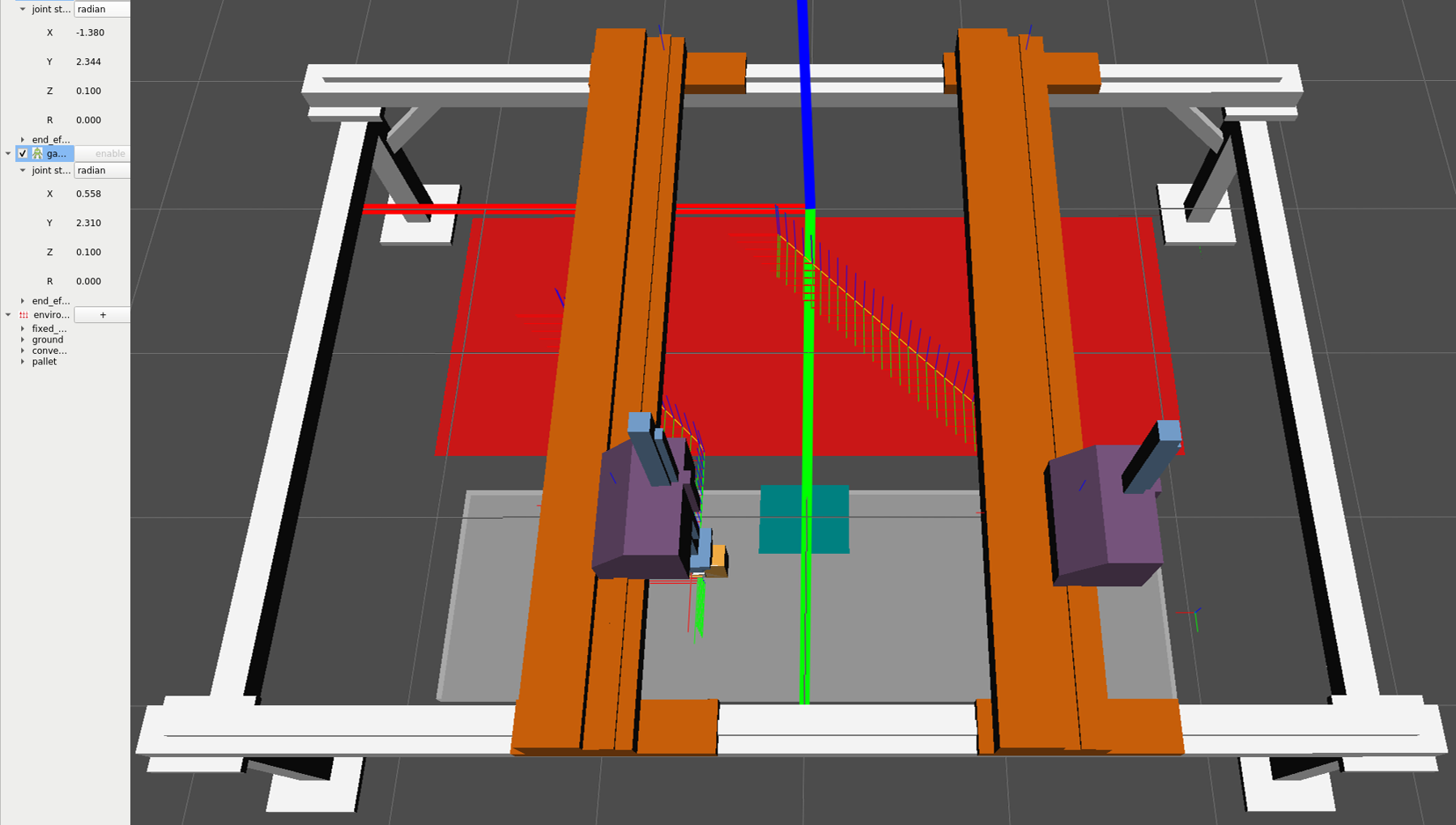}
		%
		\label{fig9:a}		
	} 
	\subfigure[Real robot]{
		\includegraphics[scale=0.03]{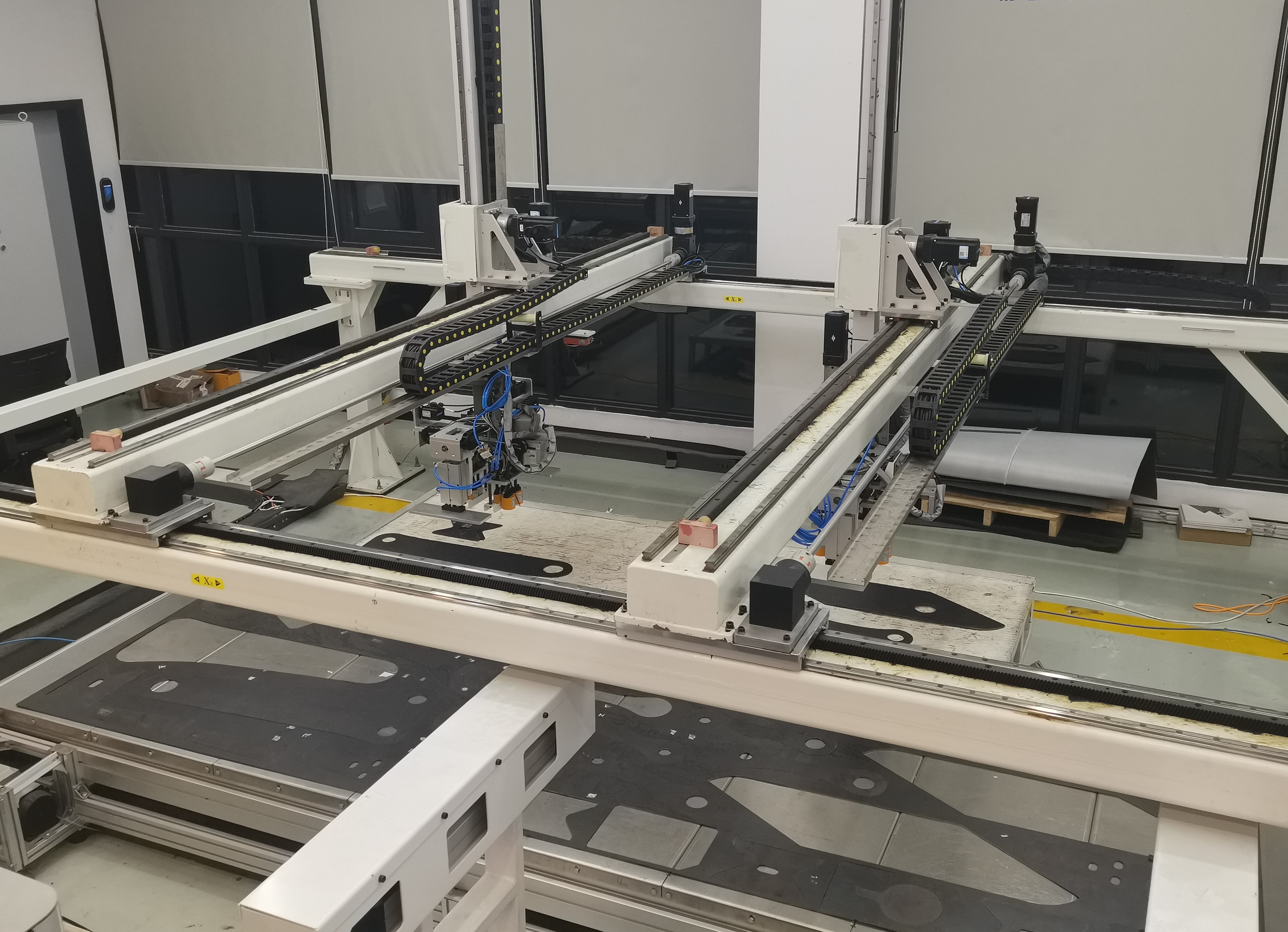}
		\label{fig9:b}
	}
	\caption{Simulator and real robot of the rearrangement system.} 
	\label{fig9}
\end{figure}

When we combine the average makespan with the average delay time proportions illustrated in Figure \ref{fig_delay}, it becomes evident that the key to enhancing efficiency in dual-arm cooperation and ultimately reducing the total makespan lies in minimizing the delay or concession time proportion. This essentially translates to maximizing the concurrent engagement of both arms in the task.

Comparing the average makespan results of our method with those of two ablations, we can find that there is a larger gap between \textit{No\_object\_encoder} and ours in both  Table \ref{tab:1} and Table \ref{tab:2}. To gain deeper insights, we present the training comparison in Figure \ref{fig4} and find that \textit{No\_object\_encoder} tends to converge earliest in both two sampling schemes. We analyze this because the modeling of object-to-object positional association substantially influences the perception of the operational cost of potential assignment pairs, which causes a more significant impact on the overall results.



\subsection{Simulator and real robot}
We have built a dual-arm rearrangement system based on ROS \cite{quigley2009ros}, which consists of a motion control simulator and a real robot as shown in Figure \ref{fig9}, in which we integrate our task assignment and dual-arm motion planning algorithm.

The pick positions of all objects are obtained in advance by a visual positioning system. Once the assigned objects are determined, the accurate pick position and target place position will be published for both arms, then the motion planning will be made in the simulator before real execution.

The demonstration of motion planning in the simulator and the comparison of rearranging fixed numbers and positions of objects using heuristics and our task assignment method in the real scene are shown in the supplemental video.

\section{Conclusion} 
This paper introduces an online reinforcement learning method that addresses the dual-arm object rearrangement problem by studying an efficient task assignment strategy. Our approach emphasizes combinatorial optimality and leverages an attention-based network to model the dependencies among different states, leading to improved performance in long-sequence decision-making. Experimental results demonstrate the effectiveness and computational efficiency of our method, especially as the number of objects increases. In future work, we plan to extend our approach to more diverse and complex multi-robot cooperation scenarios.

\section{Acknowledgement}
This work was supported in part by the NSFC (62325211, 62132021, 62372457), the National Key Research and Development Program of China (2018AAA0102200), the Major Program of Xiangjiang Laboratory (23XJ01009) and the Huxiang Youth Talent Support Program (2021RC3071). 

\bibliographystyle{IEEEtran}
\bibliography{IEEEabrv,example_paper}

\end{document}